\documentclass[journal]{IEEEtran}

\usepackage{booktabs}
\usepackage[colorlinks,
linkcolor=red, 
anchorcolor=black,
citecolor=green,
backref=page,
]{hyperref}
\usepackage{float}

\usepackage{lineno,hyperref}
\modulolinenumbers[5]
\usepackage{amsmath,amssymb,amsthm}
\usepackage{graphicx}
\usepackage[justification=centering]{subcaption}
\usepackage[numbers]{natbib}
\usepackage[ruled,linesnumbered]{algorithm2e}
\usepackage{color}
\usepackage{bbding}
\usepackage{multirow}
\usepackage{bm}
\usepackage {ulem}
\usepackage[table]{xcolor}
\usepackage[capitalize]{cleveref}
\crefname{section}{Sec.}{Secs.}
\Crefname{section}{Section}{Sections}
\Crefname{table}{Table}{Tables}
\crefname{table}{Tab.}{Tabs.}

\begin{document}
	
	\title{Single-Shot and Multi-Shot Feature Learning\\ for Multi-Object Tracking}
	
\author{Yizhe Li,
        Sanping~Zhou,~\IEEEmembership{Member,~IEEE,}
        Zheng~Qin,
        Le~Wang,~\IEEEmembership{Senior Member,~IEEE,}
        Jinjun~Wang,
        Nanning~Zheng,~\IEEEmembership{Fellow,~IEEE}
\thanks{This work was supported partly by National Key R\&D Program of China under Grant 2021YFB1714700, NSFC under Grants 62088102 and 62106192, Natural Science Foundation of Shaanxi Province under Grants 2022JC-41, China Postdoctoral Science Foundation under Grants 2020M683490 and 2022T150518, and Fundamental Research Funds for the Central Universities under Grants XTR042021005 and XTR072022001. \textit{(Corresponding author: Sanping Zhou.)}}%
\thanks{Yizhe Li, Sanping Zhou, Zheng Qin, Le Wang, Jinjun Wang and Nanning Zheng are with the National Key Laboratory of Human-Machine Hybrid Augmented Intelligence, National Engineering Research Center for Visual Information and Applications, and Institute of Artificial Intelligence and Robotics, Xi'an Jiaotong University, Xi'an, Shaanxi 710049, China.}
}

\markboth{IEEE TRANSACTIONS ON MULTIMEDIA}%
{Shell \MakeLowercase{\textit{et al.}}: A Sample Article Using IEEEtran.cls for IEEE Journals}
	
	
\maketitle
	
 \begin{abstract}
 Multi-Object Tracking~(MOT) remains a vital component of intelligent video analysis, which aims to locate targets and maintain a consistent identity for each target throughout a video sequence. Existing works usually learn a discriminative feature representation, such as motion and appearance, to associate the detections across frames, which are easily affected by mutual occlusion and background clutter in practice. In this paper, we propose a simple yet effective two-stage feature learning paradigm to jointly learn single-shot and multi-shot features for different targets, so as to achieve robust data association in the tracking process. For the detections without being associated, we design a novel single-shot feature learning module to extract discriminative features of each detection, which can efficiently associate targets between adjacent frames. For the tracklets being lost several frames, we design a novel multi-shot feature learning module to extract discriminative features of each tracklet, which can accurately refind these lost targets after a long period. Once equipped with a simple data association logic, the resulting VisualTracker can perform robust MOT based on the single-shot and multi-shot feature representations. Extensive experimental results demonstrate that our method has achieved significant improvements on MOT17 and MOT20 datasets while reaching state-of-the-art performance on DanceTrack dataset.
\end{abstract}
	
 \begin{IEEEkeywords}
 Multi-Object Tracking, Discriminative Feature Learning, Data Association.	
 \end{IEEEkeywords}
	
 \section{Introduction}
	
 \IEEEPARstart{M}{ulti-Object} Tracking~(MOT) is a fundamental task in computer vision~\cite{Wan_Cao_Zhou:2021, instance_seg}, which aims to locate targets and maintain a consistent identity for each target throughout a video sequence. As a crucial component of many applications, such as video surveillance~\cite{surveillance}, robotics~\cite{surveil} and autonomous driving~\cite{referring}, various methods are proposed to improve the performance of MOT in the past few years. In general, the existing works can be simply divided into two categories, \textit{i.e.}, tracking-by-detection~\cite{tracking-by-detection} and tracking-by-regression~\cite{tracktor}. In particular, the former ones divide MOT into two separate tasks: object detection and data association, in which a detection model is applied to detect targets in each frame, and then a data association algorithm is designed to associate detected targets of the same identity to form trajectories. What's different, the latter ones perform the object detection and data association in one step, in which they often propagate each tracklet of the previous frame to its location in the current frame.
 
\begin{figure}[t]
\centering
\includegraphics[width=1.0\linewidth]{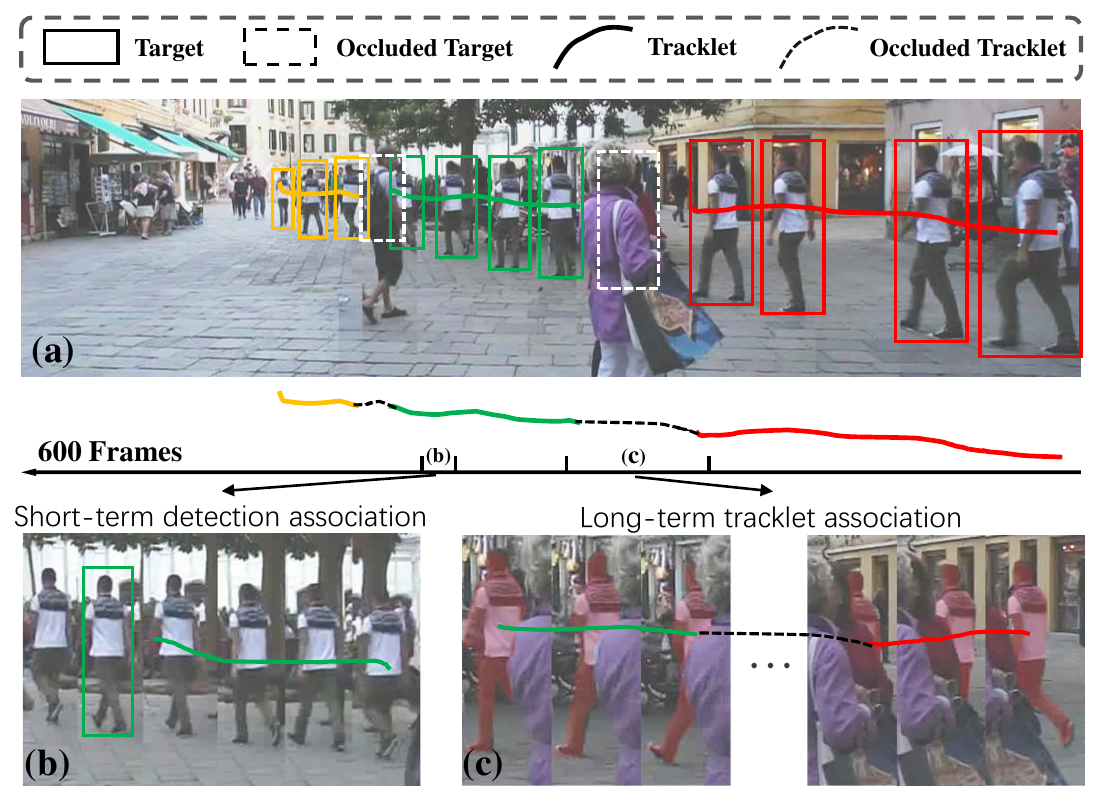}
\caption{\textbf{Observation of one target throughout a whole video sequence}, in which:~\textbf{(a)} Existing methods are severely affected by heavy occlusion and distractions in practice, which will generate tracklets in data association.~\textbf{(b)} Illustration of some normal samples. It is easy to learn the discriminative feature of each detection, so as to associate detections into tracklets. \textbf{(c)} Illustration of some occluded samples. It is better to learn the discriminative feature of each tracklet, so as to associate tracklets into trajectories.}
\label{fig:intro}
\end{figure}
 
No matter which paradigm you choose, both tracking-by-detection and tracking-by-regression methods need to overcome extreme challenges, such as mutual occlusion and background clutter, to obtain robust MOT. As shown in~\Cref{fig:intro}~(a), it is hard to keep the long-term consistency of each trajectory, which will generate a large number of tracklets and cause large identity switches in the tracking process. To overcome this problem, many pioneering works~\cite{deepsort,botsort,finetrack,qdtrack,instance_seg} introduce different feature learning models to learn discriminative feature representations for each target. For example, the recent StrongSORT~\cite{strongsort} adopts an off-the-shelf person re-identification network~\cite{bot} to extract discriminative features from input images, which is effective in associating targets across frames. What's different, some other works~\cite{jde,fairmot,relationtrack, balance} further integrate object detection and feature learning in a joint network, in which an optimal balance can be achieved in learning both fine-grained features for data association and coarse-grained features for object detection. 

Even though significant progress has been achieved in learning discriminative features for robust data association, we still argue that this problem is far from being solved in practice. There is a serious imbalance between normal samples and occluded samples, which makes it very easy for the feature learning model to overfit to normal samples. As a result, they will have a weak ability to deal with the targets with severe occlusion. To address the challenging issue, as shown in \Cref{fig:intro}~(b) and~(c), the two-stage paradigm is often applied for data association, in which the short-term data association usually aims to assign the current detection to its corresponding target in the adjacent frame, while the long-term data association often focuses on matching two adjacent tracklets after an interruption. For example, the MotionTrack~\cite{qin2023motiontrack} jointly learn short-term and long-term motion patterns to conduct robust data association in a local to global view. However, how to learn discriminative appearance features to equip with the two-stage data association process is still under exploration in the MOT community. 
	
In this paper, we propose VisualTracker, which can jointly learn single-shot and multi-shot appearance features for robust MOT. Specifically, our VisualTracker introduces two modules,~\textit{i.e.}, Single-Shot Feature Learning~(SSFL) module and Multi-Shot Feature Learning~(MSFL) module, to learn two kinds of discriminative features for short-term detection association and long-term tracklet association. To achieve the above goal, the SSFL module first takes an encoder network~\cite{attention} to conduct the pixel-level feature interaction between adjacent frames, and then aggregates the resulting feature maps to generate the discriminative features for short-term detection association. What's different, the MSFL module first utilizes a multi-head attention network~\cite{vit} to extract the frame-wise features within each tracklet, and then captures the temporal correlation via a simple fully connected layer to generate the discriminative features for long-term tracklet association. Once the short-term and long-term discriminative features are learned, a simple yet effective data association algorithm is introduced for robust MOT in complex scenarios with dense crowds and frequent occlusions. Extensive experiments on several datasets, including MOT17, MOT20 and DanceTrack, demonstrate that our VisualTracker outperforms most of the state-of-the-art methods.
 
The main contributions of this work can be summarized as follows:
\begin{itemize}
    \item We design a novel VisualTracker for robust multi-object tracking, which jointly learns single-shot and multi-shot appearance features for the two-stage data association. 
    \item We design a novel single-shot feature learning module to extract short-term discriminative features by conducting pixel-level feature interaction and aggregation.
    \item We design a novel multi-shot feature learning module to extract long-term discriminative features by enhancing the temporal correlation within each tracklet.
\end{itemize}

The rest of this paper is organized as follows: We briefly review the related work in Section~\ref{Related Works}. We present the technical details of our proposed method in Section~\ref{Our Method}. Then, extensive experiments and analysis are presented in Section~\ref{Experimental Analysis}. Finally, we conclude the paper in Section~\ref{Conclusion}.

\begin{figure*}[t]
	\centering
	\includegraphics[width=\linewidth]{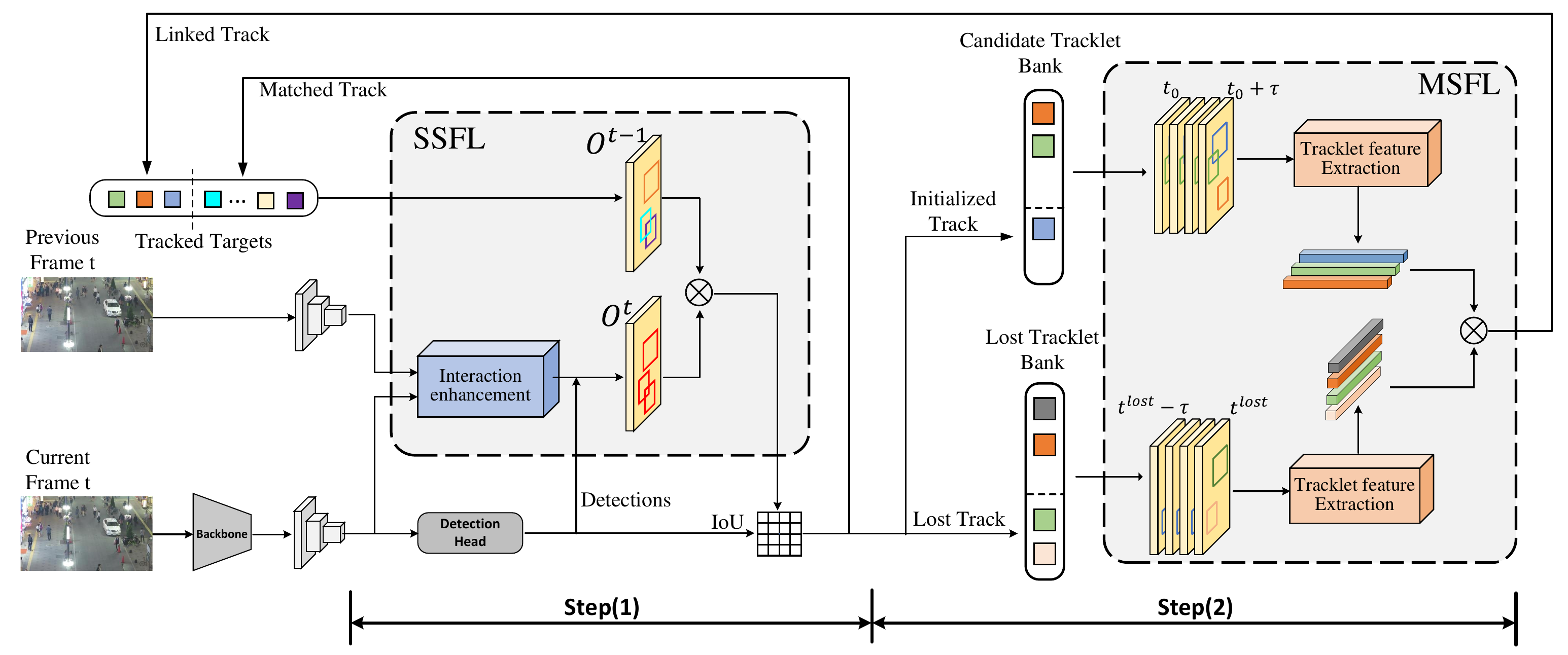}
	\caption{\textbf{Overview of proposed tracking framework}. In each frame, we first obtain the base feature pyramid from the backbone in YOLOX. After that, we perform two steps to carry out short-term and long-term association respectively. Step~(1): Single-Shot Feature Learning module enhances the base feature pyramid to obtain an ID-aware map, then we use RoIAlign~\cite{roialign} to output the target feature used in short-term association. Step~(2): Multi-Shot Feature Learning module extracts tracklet-level features for tracklets in two banks respectively to perform long-term association. Combining the association results of step~(1) and step~(2), we obtain the final tracking results.
  }
	\label{fig:framework}
\end{figure*}
	
\section{Related Work}
\label{Related Works}
\subsection{Tracking-by-Regression}
With the development of network structures and optimization techniques, many efforts have attempted to design an end-to-end framework for MOT task, thus giving rise to the tracking-by-regression paradigm. Following this paradigm, some works~\cite{tracktor,centertrack,Tubetk,chained-tracker,fft,siammot} attempt to perform object detection and location prediction in a joint network, whose challenges mainly lies in how to learn a robust mapping function from appearance to motion. For example, Tracktor~\cite{tracktor} adopts the regression head of Faster R-CNN~\cite{faster-rcnn} to regress each target bounding box across frames. In similar ways, CenterTrack~\cite{centertrack} uses pairwise frames to directly predict the target displacements for data association in a unified network. What's more, FFT~\cite{fft} further introduces optical flow to better predict the target displacements between adjacent frames, and SiamMOT~\cite{siammot} uses target patches in adjacent frames to regress bounding boxes in the next frame. Although these methods have achieved promising results, they lack the ability to model long-term dependencies across frames, thus leading to frequent identity switch during the tracking process. Different from the above methods, some other works~\cite{motr,trackformer,memot} adopt transformer-based architecture to jointly conduct object detection and data association, in which the data association is performed by updating the tracking queries while taking the new-born objects as detect queries. For example, MOTR~\cite{motr} extends the deformable DETR~\cite{deformable}, updating track queries from object queries and propagating them to the next frame as inputs of the Transformer decoder. Besides, MeMOT~\cite{memot} further builds a memory bank to store and update states of all tracked objects, which can improve the model's ability to associate long-term targets. However, we argue that the transformer-based methods are computationally intensive and not sufficiently competitive in terms of tracking performance.

\subsection{Tracking-by-Detection}
Thanks to the rapid development of object detection, various works follow the tracking-by-detection paradigm to conduct MOT. In particular, an object detector is first used to detect the location of the target in each frame. Then, data association algorithms are designed to associate the detected bounding boxes with the existing tracklets across frames. Because object detection and data association are taken as two independent tasks, this line of works mainly focus on how to conduct data association in the tracking process. On the one hand, some works~\cite{sort,bytetrack,oc-sort,Motion-aware} take the motion information of targets as a cue for data association. For example,
SORT~\cite{sort} adopts Kalman Filter~\cite{kalman1960new} to model target movement and predict the target location in the next frame, then utilizes the Hungarian algorithm~\cite{hungarian} for data association. What's different, MotionTrack~\cite{qin2023motiontrack} introduces a graph convolution network to learn the target motion pattern, and obtains more accurate offset predictions between adjacent frames. On the other hand, some works~\cite{deepsort,strongsort,fairmot,finetrack,relationtrack} introduce the appearance information to MOT, which can enhance the tracking robustness in complex scenarios with dense crowds and diverse target motion. For example, a part of these works~\cite{deepsort,strongsort} directly take an existing person re-identification network to extract the discriminative features of images in bounding boxes. Because these works take object detection and person re-identification as two independent tasks, they usually need high computational costs in practice. To address this issue, FairMOT~\cite{fairmot} implements an extra branch to learn discriminative features, which can achieve significant improvements in matching targets with similar appearance. However, this framework also poses a problem that how to achieve a balance between learning coarse-grained features for object detection and fine-grained features for person re-identification. To alleviate this issue, 
RelationTrack~\cite{relationtrack} decouples the representations used for detection and Re-ID.

However, these appearance-based MOT methods usually take the features extracted by the person re-identification~\cite{reid} model, such as~\cite{bot,zhoureid4}, to handle both short-term and long-term data association, which would lead to a weak ability in dealing with the targets with severe occlusions. To address this problem, we design a simple yet effective two-stage feature learning framework to jointly learn single-shot and multi-shot appearance features for short-term detection association and long-term tracklet association.

\section{Method}
\label{Our Method}
\subsection{Notation}
\label{Notation}
We denote the set of $M$ tracks up to frame $\mathnormal{t}$ as $\mathbb{T} = \{\mathcal{T}_{j}\}_{j=1}^M$.  $\mathcal{T}_{j}$ is a track with identity $\mathnormal{j}$ and is defined as $\mathcal{T}_{j}$ =\{$\mathbf{b}_{j}^{t_0}$, $...$, $\mathbf{b}_{j}^{t-1}$\}, where $\mathbf{b}_{j}^{t}\in\mathbb{R}^{4}$ is its bounding box at frame ${t}$, $t_0$ indicates the initialized moment of the track. The detection results of $N$ objects at frame $t$ are denoted as $\mathcal{D}^{t} = \{\mathbf{d}_{i}^{t}\}_{i=1}^N$, where $\mathbf{d}_{i}^{t}$ is the bounding box of the $i$-th detection.

At each timestamp, we take the raw image at frame $t$ as input and sequentially update the track set $\mathbb{T}$ with $\mathcal{D}^{t}$. In the tracking process, we denote the tracks not associated with any detections as lost, and take $\mathbb{T}^{\text{lost}}$ to represent them. 
For tracks initialized within the most recent few frames, we consider them as candidates for long-term associations and store them in $\mathbb{T}^{\text{cadi}}$.

\subsection{Overview}
\label{Overview}

As shown in \Cref{fig:framework}, given the current frame $t$, we adopt the backbone in YOLOX~\cite{yolox} to obtain the base feature pyramid $\{{\mathbf F}_{k}^{t}\in\mathbb{R}^{{D_k}\times{H_k}\times{W_k}}\}_{k=1}^3$, where $k$ indicates the level of the pyramid, $H_k$ and $W_k$ denote the height and width of ${\mathbf F}_{k}^{t}$, $D_k$ represents the feature dimension. The detection results $\mathcal{D}^{t}$ are acquired via the detection head of YOLOX. Subsequently, data association are conducted based on $\{{\mathbf F}_{k}^{t}\}_{k=1}^3$ in two steps: short-term detection association and long-term tracklet association. According to the short-term and long-term association results from these two steps, we update $\mathbb{T}$ from $t$ to $t-1$.

\noindent\textbf{Step(1):}
$\{{\mathbf F}_{k}^{t}\}_{k=1}^3$ along with $\{{\mathbf F}_{k}^{t-1}\}_{k=1}^3$ are fed into Single-shot Feature Learning module~(SSFL), which first performs pixel-level interaction to produce a discriminative ID-aware map $\mathbf{O}^{t}\in\mathbb{R}^{128\times H_1 \times W_1}$. Then it extracts short-term feature for each track in $\mathbb{T}$ and each detection in $\mathcal{D}^{t}$ based on $\mathbf{O}^{t-1}$ and $\mathbf{O}^{t}$ respectively. Afterward, we calculate the cosine similarity between them and obtain a similarity matrix $\mathbf{S}^\text{short}\in{\left[0, 1\right]}^{M\times N}$.
$\mathbf{S}^\text{short}$ is later fused with the IoU similarity~\cite{sort} matrix, then Hungarian algorithm\cite{hungarian} is used to achieve the short-term detection association.  


\begin{figure}[t]
    \centering
    \includegraphics[width=\linewidth]{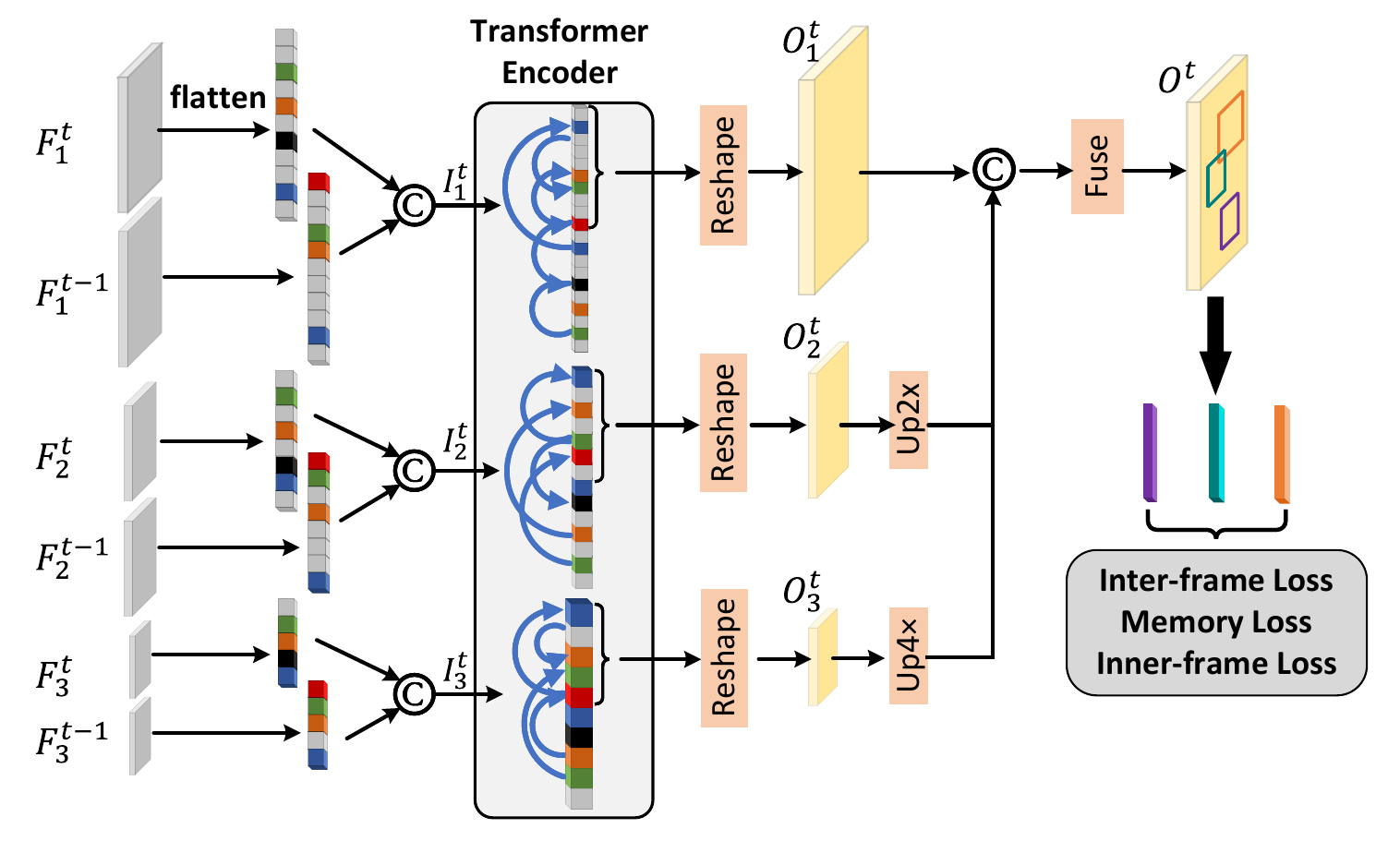}
    \caption{\textbf{Illustration of single-shot feature learning module}. We generate an ID-aware map from the base feature pyramid of adjacent frames by performing inter and inner-frame pixel-level interaction and feature aggregation.}
    \label{fig:SSFL}
\end{figure}

\noindent\textbf{Step(2):}
To refind targets that have been occluded for a long period of time, we regard tracklets initialized within the last few frames as potential candidates to associate with the lost tracklets.
For $R$ lost tracklets in $\mathbb{T}^\text{lost}$ and $U$ candidate tracklets in $\mathbb{T}^{\text{cadi}}$, Multi-shot Feature Learning module~(MSFL) extracts the tracklet-level feature for each tracklet and calculates the similarity matrix $\mathbf{S}^\text{long}\in{\left[0, 1\right]}^{R\times U}$ between them. Finally, we use the Hungarian algorithm\cite{hungarian} to determine which pair shares the same identity based on $\mathbf{S}^\text{long}$, achieving long-term tracklet association.

\subsection{Single-Shot Feature Learning module}
\label{ssfl}
To obtain more discriminative features for short-term association, we first model the pixel-level interaction between the feature pyramids of adjacent frames and aggregate the obtained feature map in each layer in interaction enhancement step. Then we extract the short-term feature and obtain the similarity matrix for association in short-term correlation construction step.



\noindent\textbf{Interaction Enhancement.} 
As shown in \Cref{fig:SSFL}, we take the feature pyramids $\{{\mathbf F}_{k}^{t-1}\}_{k=1}^3$ and $\{{\mathbf F}_{k}^{t}\}_{k=1}^3$ as input.
We first use a group of 1 $\times$ 1 convolution layers to map the channel dimension of each layer, \textit{i.e.}, ${\mathbf F}^{t-1}_{k}$ and ${\mathbf F}^{t}_{k}$, from $D_{k}$ to $D=256$. After that, we flatten the processed feature maps in space dimension and concatenate them as follows:
\begin{equation}
\begin{aligned}
{\mathbf I}_{k}^{t} = \mathcal{F}\left(\psi_k\left({\mathbf F}_{k}^{t-1}\right)\right) \oplus 
\mathcal{F}\left(\psi_k\left({\mathbf F}_{k}^{t}\right)\right),
\end{aligned}
\end{equation}
where $\oplus$ denotes concatenation operation, $\mathcal{F}(\cdot)$ represents flatten operation in space dimension, and $\psi_k(\cdot)$ denotes a group of 1 × 1 convolutional layer. $\mathbf{I}_{k}^{t}\in\mathbb{R}^{256\times{L_k}}$ is a sequence of embeddings, where $L_{k} = {H_k}{W_k}+{H_k}{W_k}$.    


We feed $\{{\mathbf I}_{k}^{t}\}_{k=1}^3$ into the transformer encoder along with their positional encoding.
To be specific, the attention mechanism~\cite{attention} captures the inner-frame and inter-frame pixel-level interaction, which enables the features of different targets to be more distinctive while the same to be consistent. Notably, We process $\{\mathbf{I}_{k}^{t}\}_{k=1}^3$ separately to avoid semantic misalignment between different levels. Subsequently, we split the output of the transformer encoder and take the half belonging to frame $t$ as $\mathbf{\hat{I}}_{k}^{t}\in\mathbb{R}^{256\times{L_k}/2}$. Then we reshape it back to the original scale (${H}_k$,${W}_k$), obtaining enhanced feature maps $\{{\mathbf O}_{k}^{t}\in\mathbb{R}^{{256}\times{H_k}\times{W_k}}\}_{k=1}^3$.

Among $\{{\mathbf O}_{k}^{t}\}_{k=1}^3$, low-level feature map contains fine-grained information such as textures, shapes, corner points, \textit{etc.}, while the high-level feature map contains semantic information. 
When scenes are crowded, occlusions and distractions will harm the semantic information, and similar appearance makes it lack discriminative ability. In this case, low-level information can serve as complementary. 
Therefore, we fuse the feature maps of different levels to enrich the target representation and obtain the ID-aware map  $\mathbf{O}^{t}\in\mathbb{R}^{128\times{H}_1\times{W}_1}$ as follows:
\begin{align}
\label{eq:fuse}
    {\mathbf O^{t}} = \psi((\delta_1({\mathbf O}_{1}^{t})\oplus\delta_2({\mathbf O}_{2}^{t})\oplus\delta_3({\mathbf O}_{3}^{t})),
\end{align}
where $\delta_k(\cdot)$ includes a upsampling operation and two Conv-ReLU~\cite{relu} layers, which is adapted to different size of ${\mathbf O}_{k}^{t}$.

\noindent\textbf{Short-term Correlation Construction.} 
 For $i$-th detection in $\mathcal{D}^{t}$ and $j$-th track in $\mathbb{T}$, we perform RoIAlign~\cite{roialign} on $\mathbf{O}^{t}$ with $\mathbf{d}_{i}^{t}$ and $\mathbf{O}^{t-1}$ with $\mathbf{b}_{j}^{t-1}$, then reshape the results to obtain short-term features as follows:

     \begin{equation}
    \begin{aligned}
            \label{eq:extract}
        {\mathbf o}^{\text{trj}}_{j} \ &= \varphi \left(\mathrm{RoIAlign}({\mathbf O}^{t-1}, \mathbf{b}_{j}^{t-1})\right),\\
        \mathbf{o}^{\text{det}}_{i} &= \varphi \left(\mathrm{RoIAlign}({\mathbf O}^t, \mathbf{d}_{i}^{t}) \right) ,
    \end{aligned}
\end{equation}
where $\varphi(\cdot)$ represents reshaping the inputs to 1D vectors and passing them through a batch normalization~\cite{bn} layer.

After that, we get features of  $M$ tracks $\{\mathbf{o}_{j}^{\text{trj}}\}_{j=1}^M$  and $N$ detections $\{\mathbf{o}_{i}^{\text{det}}\}_{i=1}^N$, then we calculate the cosine similarity $\mathbf{S}^\text{short}\in{\left[0, 1\right]}^{M\times N}$ between them as follows:
\begin{align}
        \label{eq:cos}
 {\mathbf S}^\text{short} = \{\mathbf{o}_{j}^{\text{trj}}\}_{j=1}^M\bm{\otimes}\{\mathbf{o}_{i}^{\text{det}}\}_{i=1}^N,
\end{align}
where $\bm\otimes$ denotes element-wise dot product.  


\subsection{Multi-Shot Feature Learning module}
\label{msfl}

To connect the tracklets interrupted by occlusion, we build two banks to store lost tracklets and candidate tracklets respectively. 
By constructing the long-term correlation between tracklets in two banks, we can determine which tracklet pairs share the same identity.


\noindent\textbf{Tracklet Bank.}
In a tracking scenario, when some lost targets reappear, their trajectories are often incorrectly initialized and assigned a new identity.
Therefore, we consider tracklets initialized within the last few frames as potential candidates for lost tracklets.
To implement this, we build and maintain two banks, \textit{i.e.},  the lost tracklet bank $\mathbb{T}^\text{lost}$ and the candidate tracklet bank $\mathbb{T}^\text{cadi}$.  
For $\mathbb{T}^\text{lost}$, we add lost tracklets to it and remove the tracklet when it is successfully associated or lost for more than an extended period of frames. 
For $\mathbb{T}^\text{cadi}$, we add newly initialized tracklets to it and remove the tracklet that has been alive for more than 20 frames without being associated with any lost tracklet.
Based on these two banks, we construct the long-term correlation between the two types of tracklets.

\begin{figure}[t]
    \centering
    \includegraphics[width=\linewidth]{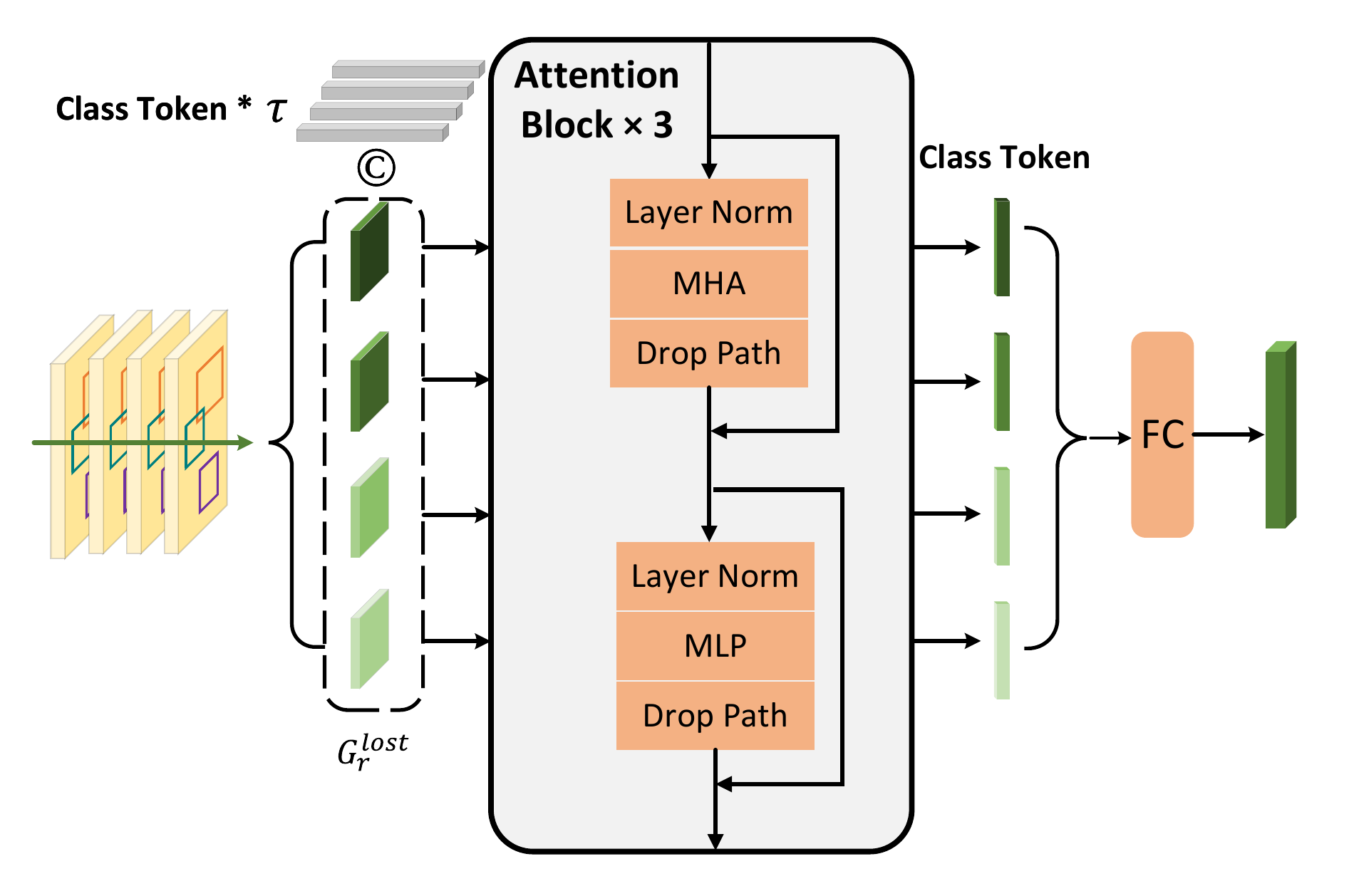}
    \caption{\textbf{ Illustration of multi-shot feature learning module}. MHA represents Multi-head Attention~\cite{attention}. We extract the tracklet-level feature for each tracklet in long-term association.}
    \label{fig:MSFL}
\end{figure}

\noindent\textbf{Long-term Correlation Construction.}
As shown in \Cref{fig:MSFL}, we first extract tracklet-level features for each tracklet from the banks described above. 
Specifically, for $r$-th tracklet in $\mathbb{T}^\text{lost}$, we perform RoIAlign with its history positions from frame $t_\varepsilon-\tau$ to frame $t_\varepsilon$ on the corresponding map in $\{\mathbf{O}^t\}_{t=t_\varepsilon-\tau}^{t_\varepsilon}$ to form $\mathbf{G}^\text{lost}_r\in \mathbb{R}^{\tau \times 128\times 4 \times 4}$, where $t_\varepsilon$ indicates the lost moment.  
Similarly, for $u$-th tracklet in $\mathbb{T}^\text{cadi}$, we acquire $\mathbf{G}^\text{cadi}_u$ based on its positions and corresponding map in $\{\mathbf{O}^t\}_{t=t_0}^{t_0 + \tau}$.


Following the spirit of ViT\cite{vit}, for each cropped tracklet feature maps, \textit{i.e.}, $\mathbf{G}^\text{lost}_r$ and $\mathbf{G}^\text{cadi}_u$, we pass them through three attention blocks separately.
Moreover, to fuse the temporal information, we adopt two independent learnable parameters, \textit{i.e.}, $\mathbf{W}^\text{lost}$ and $\mathbf{W}^\text{cadi}$, to weight the $\tau$ frame-wise features. We obtain the tracklet-level feature $\mathbf{g}_r$,  $\mathbf{g}_u \in \mathbb{R}^{128}$ as follows:
\begin{equation}
\label{eq:mlp}
    \begin{aligned}
    \mathbf{g}_r &= \mathbf{W}^\text{lost\ }  \cdot \phi(\mathbf{G}^\text{lost}_r),\\ 
    \mathbf{g}_u &= \mathbf{W}^\text{cadi} \cdot \phi(\mathbf{G}^\text{cadi}_r),
    \end{aligned}
\end{equation}
where $\phi(\cdot)$ denotes $3$ attention blocks.




Calculating the cosine similarity of the tracklet features between two banks, we obtain the matrix $\mathbf{S}^\text{long}$ used for long-term tracklet association. Each element in $\mathbf{S}^\text{long}$ represents the correlation score, which indicates whether a lost tracklet and a newly initialized tracklet belong to the same target.

\subsection{Training}
\label{Training}
\noindent{\bf Training of SSFL.}
We supervise the training process of SSFL with the total loss $\mathcal{L}^{\text{total}}$ consisting of three components computed as follows:
 
	\begin{align}
		\mathcal{L}^{\text{total}} =  \mathcal{L}^{\text{inter}} + \lambda_1\mathcal{L}^{\text{memo}} 
		+ \lambda_2\mathcal{L}^{\text{inner}},
	\end{align}
where $\mathcal{L}^{\text{inter}}$ denotes inter-frame loss, which is used to supervise target feature between adjacent frames, the memory loss $\mathcal{L}^{\text{memo}}$ is designed to ensure the temporal consistency of target representation and the purpose of inner-frame loss $\mathcal{L}^{\text{inner}}$ is to handle hard samples within the same frame. $\lambda_1$ and $\lambda_2$ are hyper-parameters for weight scaling. 

For inter-frame loss $\mathcal{L}^{\text{inter}}$, we randomly select two consecutive frames from MOT dataset as a training sample. We obtain $\mathbf{Y}^{\text{gt}}$ as ground truth, and each element of it is given by \Cref{eq:label}.


\begin{equation}
    \begin{aligned}
            \label{eq:label}
        {y}^{\text{gt}}_{ij} = 
        \begin{cases}
            1, &\mbox{if} \ {v}_{i}^{t} = {v}_{j}^{t-1} \\
            0,  &\mbox{else} 
        \end{cases}
    \end{aligned}
\end{equation}
where ${v}_{i}^{t}$ indicates the identity of the $i$-th target in frame $t$. We use cross-entropy loss to obtain $\mathcal{L}^{\text{inter}}$ based on $\mathbf{S}^\text{short}$ and $\mathbf{Y}^\text{gt}$ as follows:

\begin{align}
    \mathcal{L}^{\text{inter}} =\mathrm{CE}\left(\mathbf{S}^{\text{short}}, \mathbf{Y}^\text{gt}\right).
\end{align}

In order to maintain the temporal consistency of target representation, we design the memory loss $\mathcal{L}^{\text{memo}}$. Specifically, we store the features in $\mathcal{O}^\text{memo}$ for each target and update them recursively. For $i$-th target, we update its feature ${\mathbf{o}}^\text{memo}_i$ with a dynamic ratio factor $\alpha$ based on its current feature $\mathbf{o}_i^t$ as follows:


	

 	\begin{equation}
		\begin{aligned}
                \label{eq:memo}
            {\alpha} &= \frac{e^{{\mathbf{o}_i^t}\cdot {\mathbf{o}^\text{memo}_i}}}{\sum_{k=1}^{K} {e^{{\mathbf{o}_i^t}\cdot {\mathbf{o}^\text{memo}_k}}}},\\
		{\mathbf{o}}^\text{memo}_i &= {\alpha} \cdot {\mathbf{o}^t_i} + (1-{\alpha}) \cdot {\mathbf{o}^\text{memo}_i},
		\end{aligned}
	\end{equation}
 where $K$ indicates the number of targets in memory.
Then we calculate the memo loss as follows:

	\begin{equation}
		\begin{aligned}
			\mathcal{L}^{\text{memo}} = \sum_{i=1}^{N}\mathrm{CE}\left(\mathrm{Argmax}(\mathcal{O}^\text{memo}\bm{\otimes}\mathbf{o}_{i}^{t}), \  {v}_i\right),
		\end{aligned}
	\end{equation}
where $N$ represents the number of targets in frame $t$.


To make the features within the same frame more discriminative, we use triplet loss\cite{triplet} to calculate $\mathcal{L}^{\text{inner}}$. To be specific, we take $i$-th target in frame $t$ as the anchor, the same target in adjacent frames as positive samples. For negative samples, we select hard samples that are pretty similar to the anchor.

\noindent{\bf Training of MSFL.}
To train MSFL, we obtain the complete trajectory for each target from MOT dataset.
For each trajectory, we locate the occlusion and break it into two parts, \textit{i.e.}, front tracklet and rear tracklet.  
Then for all trajectories, we randomly select a front tracklet and a rear tracklet to form a training sample, and label positive or negative by whether they belong to the same trajectory.
We extract tracklet-level features for two tracklets in each training sample respectively as in \Cref{msfl}, then we supervise MSFL module with a cross-entropy loss as follows:




\begin{align}
        \label{eq:asso}
    \mathcal{L}^{\text{asso}} =\frac{1}{n} \sum_i^{n}-[y_{i} \log (s_{i})+(1-y_{i})\log(1-s_{i})],
\end{align}
where $s_{i}$ indicates the cosine similarity between two tracklets features in $i$-th sample. $y_{i}$ is the ground truth label, in which 1 and 0 represent whether the two tracklets belong to the same target or not respectively.

\begin{table}[t]
    \centering  
    \caption{Comparison with the state-of-the-art methods on the DanceTrack\cite{dancetrack} test set. The two best results for each metric are highlighted in red and blue. Our method shares detections with our baseline ByteTrack and is highlighted in gray.}
    \label{tab:dance}
    \resizebox{0.95\linewidth}{!}{
        \setlength{\tabcolsep}{4.8pt}
\begin{tabular}{l|ccccc}
\hline
\textbf{\textbf{Methods}}                        & \textbf{\textbf{HOTA$\uparrow$}} & \textbf{\textbf{IDF1$\uparrow$}} & \textbf{\textbf{MOTA$\uparrow$}} & \textbf{\textbf{DetA$\uparrow$}} & \textbf{\textbf{AssA$\uparrow$}} \\ \hline
\textcolor{red}{\textbf{$motion$ :}}\\
ByteTrack \cite{bytetrack}                       & 47.7                             & 53.9                             & 89.6                             & 71.0                             & 32.1                             \\
MotionTrack \cite{qin2023motiontrack}            & 48.9                             & 44.3                             & 91.1                             & \textcolor{red}{\bf82.3}                             & 29.2                             \\
OC-SORT \cite{oc-sort}                           & \textcolor{blue}{\textbf{55.1}}  & 54.6                             & \textcolor{red}{\bf92.0}         & {80.3}  & 38.3                             \\

\midrule
\textcolor{red}{\textbf{$regression$ :}}\\
CenterTrack \cite{centertrack}                   & 41.8                             & 35.7                             & 86.8                             & 78.1                             & 22.6                             \\ 
TraDes \cite{Track-to-detect}                    & 43.3                             & 41.2                             & 86.2                             & 74.5                             & 25.4                             \\
TransTrack \cite{transtrack}                     & 45.5                             & 45.2                             & 88.4                             & 75.9                             & 27.5                             \\
GTR \cite{gtr}                           & 48.0                             & 50.3                             & 84.7                             & 72.5                             & 31.9                             \\
MOTR \cite{motr}                                 & 54.2                             & 51.5                             & 79.7                             & 73.5                             & \textcolor{red}{\textbf{40.2}}   \\
\midrule
\textcolor{red}{\textbf{$embedding$ :}}\\
FairMOT \cite{fairmot}                           & 39.7                             & 40.8                             & 82.2                             & 66.7                             & 23.8                             \\
QDTrack \cite{qdtrack}                           & 45.7                             & 44.8                             & 83.0                             & 72.1                             & 29.2                             \\ 
DeepSORT~\cite{deepsort}                         & 45.6                             & 47.9                             & 87.8                             & 71.0                             & 29.7                             \\
FineTrack\cite{finetrack}                        & 52.7                             & \textcolor{red}{\textbf{59.8}}   & {89.9}                           & 72.4                             & {38.5}                           \\
\rowcolor{gray!25}            \bf{VisualTracker} & \textcolor{red}{\bf56.7}         & \textcolor{blue}{\bf58.2}        & \textcolor{blue}{\bf91.2}        & \textcolor{blue}{\bf80.6}         & \textcolor{blue}{\bf40.0} 
\\\bottomrule
\end{tabular}
    }
    \label{table1}  
\end{table}

\section{Experiments}
\label{Experimental Analysis}
\subsection{Settings}
\noindent\textbf{Datasets.}
	We evaluate our VisualTracker on MOT17~\cite{mot16},  MOT20~\cite{mot20} and DanceTrack~\cite{dancetrack} datasets. The experiments conducted on MOT17 and MOT20 are under the “private detection” protocol.
    MOT17 consists of 7 sequences for training and 7 sequences for testing. MOT20 is a dataset of highly crowded scenes, with 4 sequences for training and 4 sequences for testing. 
    Since the MOT17 and MOT20 do not provide a validation set, we divide the training set, where the first half is used to train SSFL and MSFL while the second half serves as the validation set.
    DanceTrack is a multi-human tracking dataset in dancing scenes. It provides 40, 25, and 35 videos as training, validation, and test sets. Targets in each sequence have similar appearance and diverse motions, and suffer from severe occlusions and crossovers, which poses a challenge for data association.

\noindent\textbf{Metrics.}
We adopt CLEAR-MOT~\cite{idf1} metrics containing MOTA, IDF1, IDs, FP, FN, \textit{etc.}, as well as HOTA, DetA and AssA which are proposed in ~\cite{hota} to evaluate different aspects of tracking performance.
In particular, MOTA is computed based on FP, FN, and IDs, which focuses on localization performance, and IDF1 emphasizes association performance. Compared with them, HOTA takes localization accuracy into account, and comprehensively balances detection, association, and localization effects. 
	
\noindent\textbf{Implementation Details.}
We adopt YOLOX as our detector, following YOLOX settings in ByteTrack.
For SSFL, the dimension of the short-term feature is set to 2048, the size of the ID-aware map is $128 \times 100 \times 180$ for MOT17 and DanceTrack, $128 \times 112 \times 200$ for MOT20. For MSFL, the dimension of the long-term feature is set to 128, length of the time window $\tau$ for extracting tracklet-level features is set to $4$. For the lost tracklets in $\mathbb{T}^\text{lost}$, we keep them for 30 frames in DanceTrack and 100 frames in MOT17 and MOT20.
During training, hyper-parameters for weight scaling $\lambda_1$ and $\lambda_2$ are set to 0.2 and 1.0 respectively. 

\begin{table}[t]
    \centering  
    \caption{Comparison with the state-of-the-art methods on the MOT17 test set. The two best results for each metric are highlighted in red and blue. Our method shares detections with our baseline ByteTrack and is highlighted in gray.}
     \label{tab:17}
    \resizebox{0.95\linewidth}{!}{
        \setlength{\tabcolsep}{1.3pt}
        \begin{tabular}{l|ccccccc}
            \toprule
            \textbf{Methods}  & \textbf{MOTA$\uparrow$} & \textbf{IDF1$\uparrow$} & \textbf{HOTA$\uparrow$} & \textbf{FP(${10^3}$)$\downarrow$}  & \textbf{FN(${10^3}$)$\downarrow$}  & \textbf{IDs$\downarrow$}   & \textbf{Frag$\downarrow$}\\
            \hline
            \textcolor{red}{\textbf{$motion$ :}}\\
            ByteTrack~\cite{bytetrack}   & 80.3            & 77.3             & 63.1                & 25.5            & \textcolor{blue}{\bf83.7}            & 2196              & 2277    \\
            OC-SORT~\cite{oc-sort}        & 78.0       & 77.5      & 63.2        & \textcolor{red}{\bf15.1}      & 108.0      & 1950     & 2040\\
            MotionTrack~\cite{qin2023motiontrack} & \textcolor{red}{\bf81.1}  & \textcolor{red}{\bf80.1}  & \textcolor{red}{\bf65.1}  &23.8 & \textcolor{red}{\bf81.7}  & \textcolor{blue}{\bf1140}  & \textcolor{blue}{\bf1605}   \\

                      

            \midrule
            \textcolor{red}{\textbf{$regression$ :}}\\
            
            
            TransTrack\cite{transtrack}      & 74.5       & 63.9      & 43.9
            & 28.3     & 112.1      & 3663   & -   \\
            MOTR~\cite{motr}  & 73.4            & 68.6             & 57.8           & -           & -            & 2439             & -            \\
            MeMOT~\cite{memot}  & 72.5            & 69.0             & 56.9                   & 37.2            & 115.2           & 2724              & -                  \\

            
           
            \midrule
            \textcolor{red}{\textbf{$embedding$ :}}\\
            QDTrack~\cite{qdtrack}  & 68.7            & 66.3             & 53.9               & 26.6            & 146.6           & 3378              & 8091               \\
            SOTMOT~\cite{sotmot}  & 71.0            & 71.9             & -                       & 39.5            & 119.0           & 5184              & -                  \\
            Semi-TCL~\cite{semi} & 73.3            & 73.2             & 59.8                  & 22.9            & 125.0           & 2790              & 8010               \\
            SiamMOT~\cite{siammot}  & 76.3            & 72.3             & -                     & -                & -                & -                 & -               \\
            
            CSTrack~\cite{cstrack}   & 74.9            & 72.6             & 59.3                & 23.8            & 114.3           & 3567              & 7668               \\

            MTrack~\cite{Towards-Discriminative} & 72.1            & 73.5             & -                       & 53.4            & 101.8           & 2028              & -                  \\
            FairMOT~\cite{fairmot}  & 73.7            & 72.3             & 59.3            & 27.5            & 117.5           & 3303              & 8073               \\
            RelationTrack~\cite{relationtrack}   & 73.8            & 74.7             & 61.0                      & 28.0            & 118.6           & 1374              & 2166               \\
            ReMOT~\cite{remot} & 77.0      & 72.0        & 59.7           & 33.2     & 93.6     & 2853      & 5304\\
            GHOST\cite{simplecue}           & 78.7       & 77.1          & -         & -      & -      & 2325      & - \\
            FineTrack\cite{finetrack}         & {80.0}       & {79.5}      & {64.3}       & \textcolor{blue}{\bf21.8}      & {90.1}      & {1272}     & {1839}     \\
            \rowcolor{gray!25}
            \bf{VisualTracker}      & \textcolor{blue}{\bf80.6}     & \textcolor{blue}{\bf79.6}     & \textcolor{blue}{\bf64.5}  
            & 21.9     & {86.6}     & \textcolor{red}{\bf1092}     & \textcolor{red}{\bf1539}
            \\\bottomrule
        \end{tabular}
    }
\end{table}

\begin{table}[t]
    \centering  
    \caption{Comparison with the state-of-the-art methods on the MOT20 test set. The two best results for each metric are highlighted in red and blue. Our method shares detections with our baseline ByteTrack and is highlighted in gray.}
     \label{tab:20}
    \resizebox{0.95\linewidth}{!}{
        \setlength{\tabcolsep}{1.3pt}
        \begin{tabular}{l|ccccccc}
            \toprule
            \textbf{Methods} & \textbf{MOTA$\uparrow$} & \textbf{IDF1$\uparrow$} & \textbf{HOTA$\uparrow$} & \textbf{FP(${10^3}$)$\downarrow$}  & \textbf{FN(${10^3}$)$\downarrow$}  & \textbf{IDs$\downarrow$}   & \textbf{Frag$\downarrow$}\\ 
            \hline  
            \textcolor{red}{\textbf{$motion$ :}}\\
            ByteTrack~\cite{bytetrack} & 77.8  & 75.2       & 61.3                   & 26.2            &\textcolor{blue}{\bf87.6}            & 1223              & 1460 \\
            OC-SORT~\cite{oc-sort}  & 75.5  & 75.9      & 62.1                 & \textcolor{red}{\bf18.0}            & 108.0            &\textcolor{red}{\bf913}              & \textcolor{red}{\bf1198}\\
            MotionTrack~\cite{qin2023motiontrack}   &\textcolor{red}{\bf78.0}   & {76.5}                   & {62.8}        & 28.6           &\textcolor{red}{\bf84.2}   & {1165}     & {1321} \\
            
            \midrule
            \textcolor{red}{\textbf{$regression$ :}}\\
            Tracktor++~\cite{tracktor}  & 52.6  & 52.7         & 42.1               & -            & -          & 1648              & - \\
            TransTrack~\cite{transtrack}  & 65.0  & 59.4         & 48.5               & 27.2            & 150.2           & 3608              &-   \\
            MeMOT~\cite{memot}   & 63.7  & 66.1           & 54.1                       & 47.9            & 138.0           & 1938              & -                  \\
            \midrule
            \textcolor{red}{\textbf{$embedding$ :}}\\
            FairMOT~\cite{fairmot}  & 61.8    & 67.3         & 54.6                     & 103.4           & 88.9            & 5243              & 7874 \\
            Semi-TCL~\cite{semi}  & 65.2    & 70.1         & 55.3                     & 61.2           & 115.0            & 4139              & 8508 \\
            
            CSTrack~\cite{cstrack}   & 66.6  & 68.6           & 54.0                  & 25.4            & 144.4           & 3196              & 7632 \\
            SiamMOT~\cite{siammot}  & 67.1  & 69.1           & -                           & -                & -                & -                 & - \\
            RelationTrack~\cite{relationtrack} & 67.2  & 70.5  & 56.5
            & 61.1  & 104.6  & 4243  & 8236\\
            SOTMOT~\cite{sotmot}     & 68.6        & 71.4       & 57.4                       & 57.1            & 101.2           & 4209              & 7568 \\
            MTrack~\cite{mtrack}  & 63.5  & 69.2           & -                           & 96.1            & {87.0}            & 6031              & -\\
            FineTrack~\cite{finetrack}  &\textcolor{blue}{\textbf{77.9}}  & \textcolor{red}{\textbf{79.0}}  &\textcolor{red}{\textbf{63.6}}  &{24.4}  & 89.0 &\textcolor{blue}{\textbf{980}} &{1406}\\  
            \rowcolor{gray!25}
            \bf{VisualTracker}    &\textcolor{red}{\textbf{78.0}}  &\textcolor{blue}{\textbf{77.4}}      &\textcolor{blue}{\textbf{63.4}}       &\textcolor{blue}{\textbf{24.0}} 
            &{88.9}      &{1093}     &\textcolor{blue}{\textbf{1216}}
            \\\bottomrule
        \end{tabular}
    }
\end{table}



\subsection{Comparison with the State-of-the-Art Methods}
In this part, we compare the performance of VisualTracker with previous methods on MOT17, MOT20 and DanceTrack benchmark datasets. Results reported in this part are directly obtained from the official test server of MOT Challenge and DanceTrack competition website. Different from some appearance-based methods which introduce extra data to train an existing Re-ID model for better identity embeddings, our method only uses the MOT dataset for training and does not utilize any additional annotations for supervision.

\begin{table}[t]
    \centering  
	\caption{Ablation studies on Single-Shot Feature Learning module~(S) and Multi-Shot Feature Learning module~(M) of VisualTracker on the DanceTrack validation set.}
        \label{tab:base_ablation}
        \renewcommand{\arraystretch}{1.2}
		\resizebox{0.95\linewidth}{!}{
			\setlength{\tabcolsep}{0.2em}
			\begin{tabular}{l|cccccc}
				\toprule
				Setting & IDF1 $\uparrow$ & HOTA $\uparrow$ & MOTA $\uparrow$& DetA $\uparrow$ & AssA $\uparrow$    &IDs $\downarrow$\\ 
				\hline
				Baseline  & 51.3 & 47.1 & 88.5  & 71.2   & 31.3    & 761
				\\
				Baseline+S & 52.4 & 53.0 & 90.0  & \bf79.5   & 35.5         & 741\\

				Baseline+S+M  & \bf54.2 & \bf53.2  & \bf90.1 & 79.3   & \bf36.0    & \bf601
				\\\bottomrule
			\end{tabular}
		}
		\label{table3}  
	\end{table}
 
\begin{table}[t]
 \centering    
    \caption{Component-wise analysis of SSFL on MOT17 validation set. Fusion represents the feature map aggregation described in \Cref{eq:fuse}, $\mathcal{L}^\text{memo}$ is memory loss and $\mathcal{L}^\text{inner}$ is inner-frame loss.}
        \label{tab:ab_ssfl}
		\resizebox{0.95\linewidth}{!}{
                \setlength{\tabcolsep}{0.9em}
			\begin{tabular}{c|ccc|ccc}
				\toprule
				Settings & Fusion & $\mathcal{L}^\text{memo}$ & $\mathcal{L}^\text{inner}$  & MOTA$\uparrow$ & IDF1$\uparrow$  \\ \hline
				1&$\checkmark $  &             &                                &66.6     &60.5            \\
				2  &$\checkmark $ &$\checkmark $             &                                &70.5      &64.2            \\
				3&  &$\checkmark $             &$\checkmark $                                &71.2      &63.5            \\ \hline
				SSFL&$\checkmark $             &$\checkmark $               &$\checkmark $                &\bf71.8     &\bf66.5          \\ \bottomrule
			
   \end{tabular}
		}
	\end{table}

 \begin{table}[!ht]
\caption{Tracking performance comparison of SSFL and other existing Re-ID methods on MOT17 validation set. The first row represents identity embedding from the YOLOX backbone. The second and third rows represent identity embedding extracted through existing Re-ID models and the last row is our SSFL module.}
 \centering
 \label{comparison}
 \resizebox{0.95\linewidth}{!}{
        \setlength{\tabcolsep}{5.8pt}
\begin{tabular}{c|cccccc}
\toprule
Model                                   &{MOTA$\uparrow$} & {IDF1$\uparrow$} & {MT$\uparrow$} & {ML$\downarrow$} & {FP$\downarrow$} & {FN$\downarrow$} \\ \hline

Base & 61.0                     & 57.3                     & 132                    & 49                     & 6284                   & 13217                  \\ \hline

BoT                                                      & 70.5                     & 66.0            & 183                    & 44                     & 4484                   & 10925                  \\ 

SBS                                                      & 71.0                     & 65.6            & 181                    & 46                     & 4342                   & 10791                  \\ 

\textbf{SSFL}                                            & \bf{71.8}            & \bf{66.5}                     & \bf{186}           & \bf{42}            & \bf{4176}          & \bf{10497}         \\ \bottomrule
\end{tabular}
}
\end{table}

\noindent{\bf DanceTrack.} \Cref{tab:dance} shows the comparison of our proposed method with existing methods on the test set of DanceTrack, which features the similar appearance and diverse motions. 
With the same detection results, our VisualTracker achieves significant improvements compared to the baseline with a gain of \textbf{+9.0\%} HOTA, \textbf{+4.3\%} IDF1, \textbf{+1.6\%} MOTA,  \textbf{+9.6\%} DetA and \textbf{+7.9\%} AssA. Meanwhile, our method also achieves the best performance among the embedding-based methods. 
It is worth noting that similar appearance in DanceTrack makes embedding-based methods perform poorly, VisualTracker still yields much better performance and the highest HOTA, which indicates the superiority of our method.


\noindent{\bf MOT17.} Targets in MOT17 have relatively small and linear motions, these characteristics lead to the high performance of motion-based methods. As shown in \Cref{tab:17}, VisualTracker still achieves the best results on the MOT17 benchmark for most key metrics among embedding-based methods~(\textit{i.e.}, 80.6\% MOTA, 79.6\% IDF1, 64.5\% HOTA, etc.). Our SSFL focuses on learning more discriminative features for normal samples, high IDF1 (79.6\%) and AssA (64.5\%) indicate the effectiveness of SSFL in short-term detection association. It is worth mentioning that VisualTracker achieves the lowest IDs(1092) and Frag(1539) among all methods because MSFL successfully refinds lost targets in long-term association, which indicates the effectiveness of MSFL in long-term tracklet association.


\begin{figure}[t]
    \centering
    \includegraphics[width=0.98\linewidth]{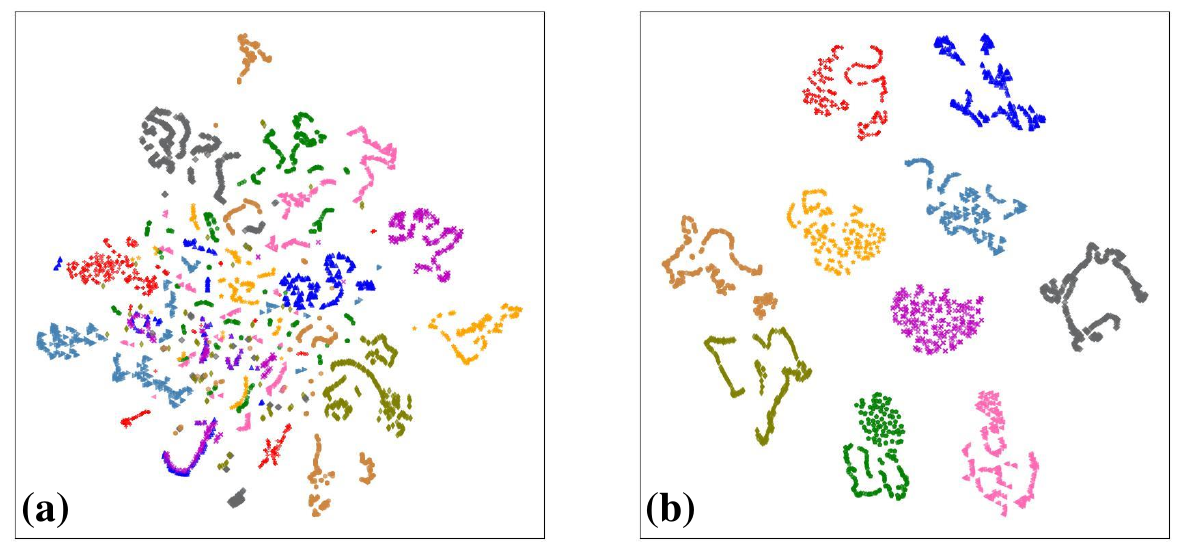}
    \caption{\textbf{Visualization of targets' features in MOT17-02}. \textbf{(a)}: Short-term features sampled from the feature pyramid of YOLOX backbone. \textbf{(b)}: Short-term features produced by SSFL. The points in the same color represent embeddings of one target at different frames.}
    \label{fig:visual_emb}
\end{figure}

\begin{figure}[t]
    \centering
    \includegraphics[width=0.98\linewidth]{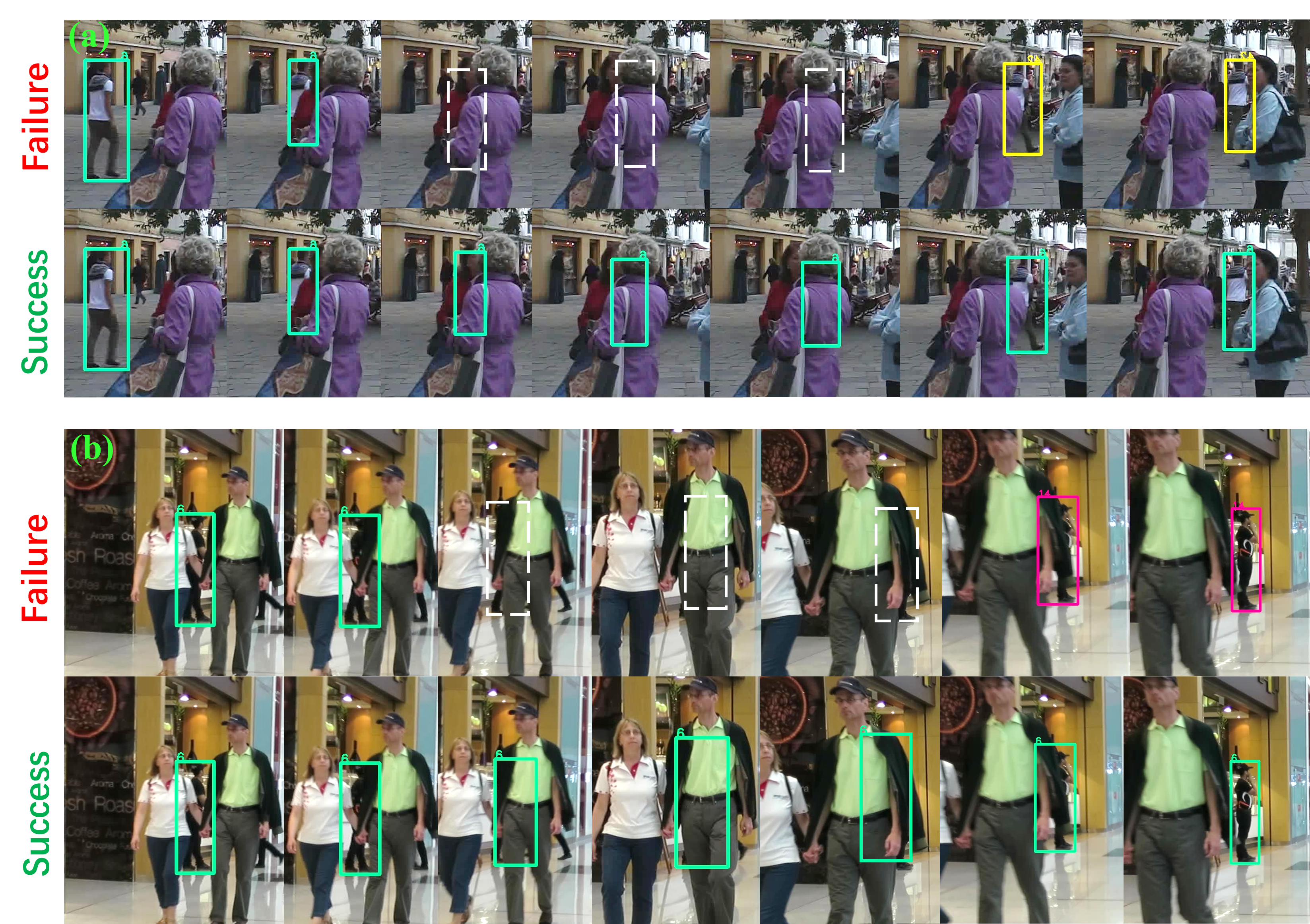}
    
    \caption{\textbf{Visualization of long-term tracklet association}. For a clear demonstration, images in~(a) are flipped horizontally. The box color corresponds to the ID and the dashed boxes represent the occluded target.}
    \label{fig:refind}
    \vspace{-0.5cm}
\end{figure}

\noindent{\bf MOT20.} Compared with MOT17, MOT20 features denser crowds and more frequent occlusions, which poses a challenge for appearance-based methods. As shown in \Cref{tab:20}, our VisualTracker still achieves a comparable performance with SOTA, outperforming ByteTrack in almost every key metric.
Note that we use exactly the same hyper-parameters as in MOT17, which implies the generalization capability of our method. The result shows the robustness of the two-stage feature learning strategy when handling complex scenarios with dense crowds and occlusion.

Note that MotionTrack is specifically designed for pedestrian scenarios, 
we still achieve comparable performance with it on MOT17\&MOT20 datasets.
On DanceTrack dataset with similar appearance and diverse motions, our method outperforms MotionTrack by 7.8 $\%$ on HOTA, 13.9\% on IDF1 and 10.2\% on IDF1 with superior identity embeddings.

\begin{figure*}[t]
    \centering
    \includegraphics[width=0.85\linewidth]{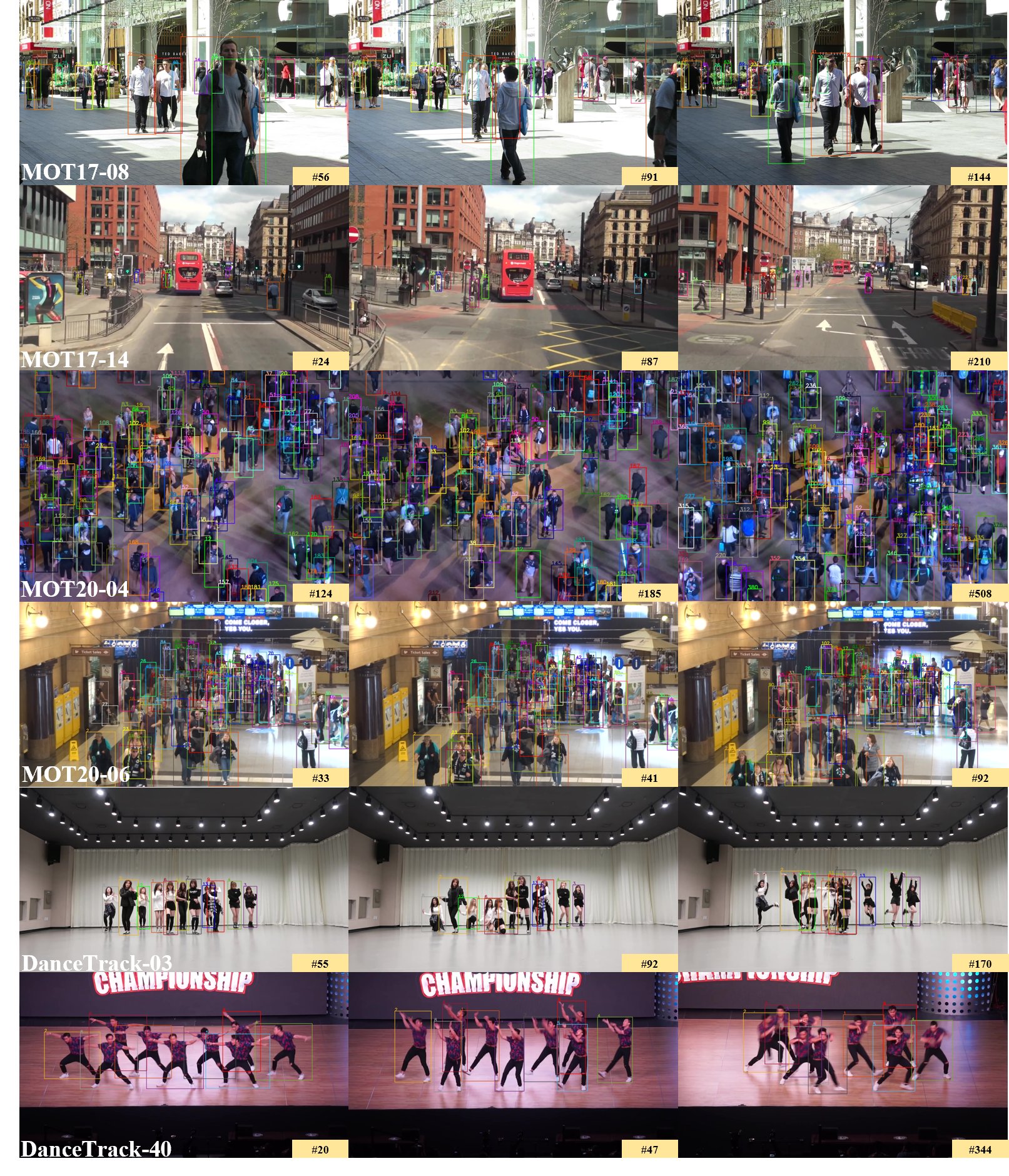}
    \caption{\textbf{Tracking results visualization of VisualTracker on the test sets of MOT17, MOT20 and DanceTrack}.}
    \label{fig:tracking_res}
\end{figure*}
 
\subsection{Ablation Study}
In this section, we verify the effectiveness of VisualTracker through ablation studies. 

\noindent {\bf Effect of SSFL and MSFL.}
We first conduct ablation experiments to verify the effectiveness of each main component of VisualTracker, \textit{i.e.}, SSFL and MSFL. We follow the same experiment settings with our baseline ByteTrack to ensure fairness and reliability. As shown in \Cref{tab:base_ablation}, SSFL significantly improves IDF1, HOTA, MOTA and IDS, indicating the effectiveness of discriminative short-term features. MSFL achieves improvement on IDF1, and AssA and reduces IDs by a considerable margin. This is because MSFL successfully associates some lost tracklets with newly initialized tracklets, indicating the effectiveness of tracklet-level features in long-term tracklet association.

\noindent {\bf Component-wise Analysis within SSFL.}
We conduct component-wise ablative experiments on MOT17 validation set to further analyze the effectiveness of each component in SSFL when tracking. As shown in \Cref{tab:ab_ssfl}, the introduction of memory loss brings significant performance gain~(3.9\% in MOTA, 3.7\% in IDF1), inner-frame loss focuses on hard samples within the same frame in tracking scenario, which increases MOTA by 1.3\% and IDF1 by 2.3\%. Feature fusion enriches the target representation, improving 0.4\% in MOTA and 2.6\% in IDF1. Incorporating the components above, we get the complete SSFL. The ablative experiments prove the effectiveness of several designed components, demonstrating the value of exploring more discriminative target representation.

\noindent {\bf Comparison of SSFL with Other Appearance Models. }
In this part, we use different appearance models, \textit{i.e.}, YOLOX backbone, two off-the-shelf Re-ID networks, and our proposed SSFL to obtain identity embedding.
Considering there is a gap between Re-ID and MOT tasks~\cite{simplecue}, the model's performance on Re-ID metrics does not sufficiently represent its performance in a tracking scenario.
Therefore, we use the metrics of MOT task to measure the performance of each method. To be specific, we replace the common IOU metric in association step with pure feature similarity during inference and evaluate these methods with tracking metrics. 
As shown in \Cref{comparison}, our method has significant advantages over embeddings from YOLOX backbone~(10.8$\%$ on MOTA, 9.2$\%$ on IDF1). 
Compared with existing Re-ID models, SSFL outperforms these models on most metrics, indicating that our SSFL is more effective for feature learning in MOT with a smaller computational overhead.
It's worth noting that BoT and SBS are pre-trained on the Re-ID dataset and fine-tuned on the MOT dataset, while our method only uses MOT dataset for training and does not use any additional labels for supervision.

\subsection{Visualization}
\noindent {\bf Visualization of Short-term Features.}
We visualize the targets' identity embeddings with and without SSFL based on the t-SNE algorithm in~\Cref{fig:visual_emb}. 
The identity embeddings without SSFL are sampled from the base feature pyramid of YOLOX backbone and the identity embeddings with SSFL are extracted as described in~\Cref{ssfl}.
As shown in~\Cref{fig:visual_emb}, the identity embeddings produced by SSFL are more discriminative, which means that the embeddings of the same target at different frames are well clustered and the embeddings of different targets are clearly distinguished.
The visualization result demonstrates that the proposed SSFL can effectively improve the distinguishability of target features for short-term detection association.

\noindent {\bf Visualization of Long-term Tracklet Association.}
As shown in \Cref{fig:refind}, in the previous methods, target trajectories are often incorrectly initialized and assigned a new identity after a long-term occlusion. MSFL takes features in multiple frames to produce a discriminative tracklet-level feature for long-term tracklet association. Therefore, our method is able to identify the lost target as soon as it reappears from long-term occlusion thus forming a complete trajectory.

\noindent {\bf Visualization of Tracking Results.}
We visualize several tracking results on the test sets of MOT17, MOT20 and DanceTrack in~\Cref{fig:tracking_res}, the results of MOT17-08 and MOT17-14 show that our VisualTracker performs well in scenarios with frequent target distractions and camera movement.
The results of MOT20-04 and MOT20-06 show the sound tracking performance in scenarios with dense crowds and frequent occlusions.
The results of DanceTrack-03 and DanceTrack-40 show that in scenarios with diverse motion patterns and similar appearance, our VisualTracker is still able to achieve a satisfying tracking performance.
In a word, the results prove that VisualTracker can achieve robust and accurate tracking performance even under challenging conditions.

\section{Conclusion}
\label{Conclusion}
In this paper, we have argued that there are two different types of association in the MOT task and analyzed the necessity of learning specific features for these two kinds of data association.
Based on this, we propose VisualTracker, which follows a two-stage feature learning paradigm to jointly learn single-shot and multi-shot features for
different kinds of targets.
Correspondingly, the single-shot feature learning module extracts discriminative features of each detection and associates targets between adjacent frames, while the multi-shot feature learning module extracts discriminative features of each tracklet, which can accurately refind lost targets after a long period.
The effectiveness of single-shot and multi-shot feature learning paradigm has been verified through ablation experiments. The experiment results also demonstrate that the proposed framework achieves significant improvement and reaches state-of-the-art performance on multiple datasets.
We hope this work can provide a new paradigm and solution for feature learning in MOT.

\bibliographystyle{splncs04}
\bibliography{reference}

\newpage
\begin{IEEEbiography}[{\includegraphics[width=1.0in,height=1.2in,clip]{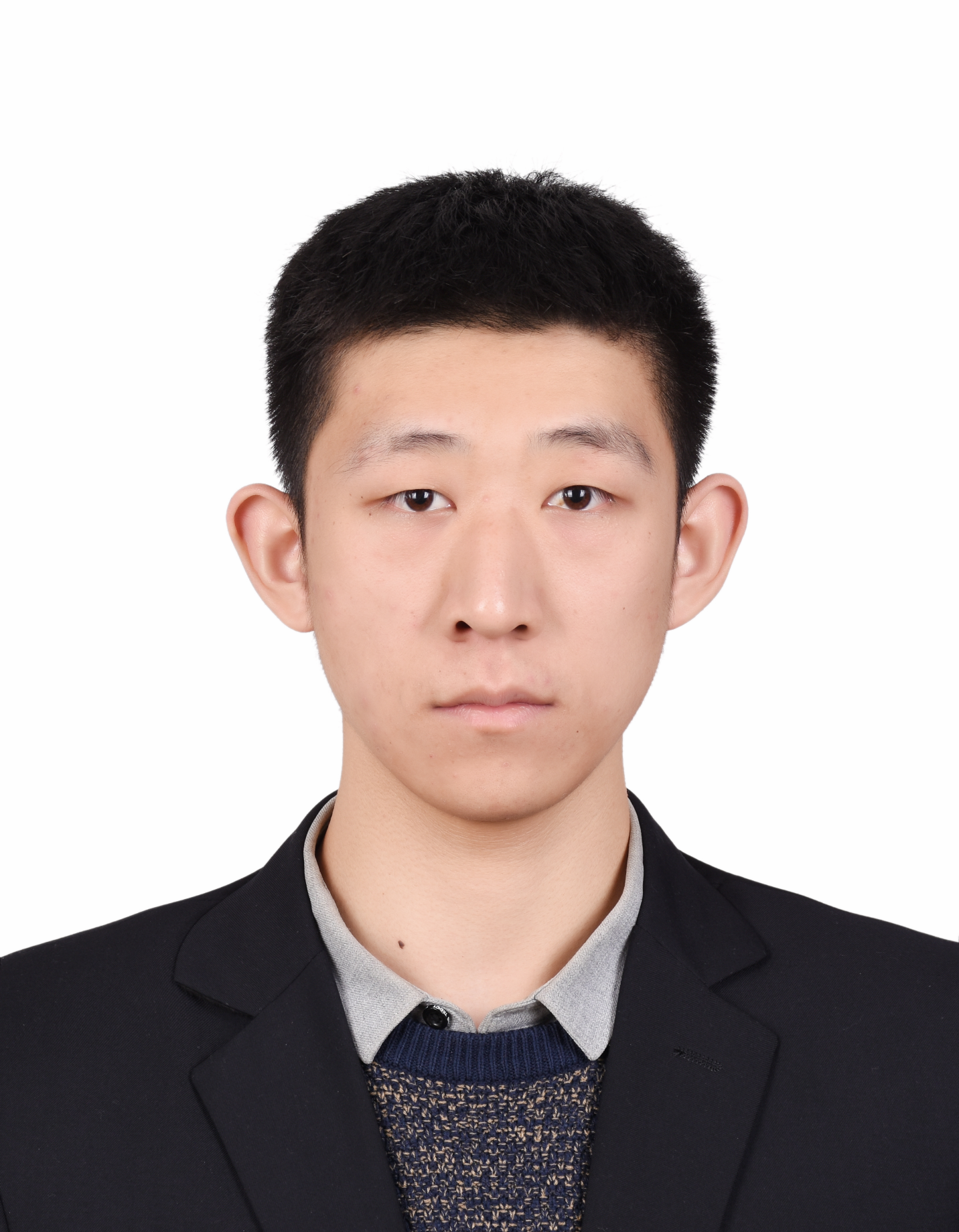}}]{Yizhe Li}
received the B.S. degree in control science and
engineering from the Xi'an Jiaotong University, Xi'an, China, in 2022. He is currently working toward the M.S. degree in artificial intelligence from Xi'an Jiaotong University. His research interests include computer vision and multi-object tracking.
\end{IEEEbiography}
\vspace{-30pt} 
\begin{IEEEbiography}[{\includegraphics[width=1.0in,height=1.2in,clip]{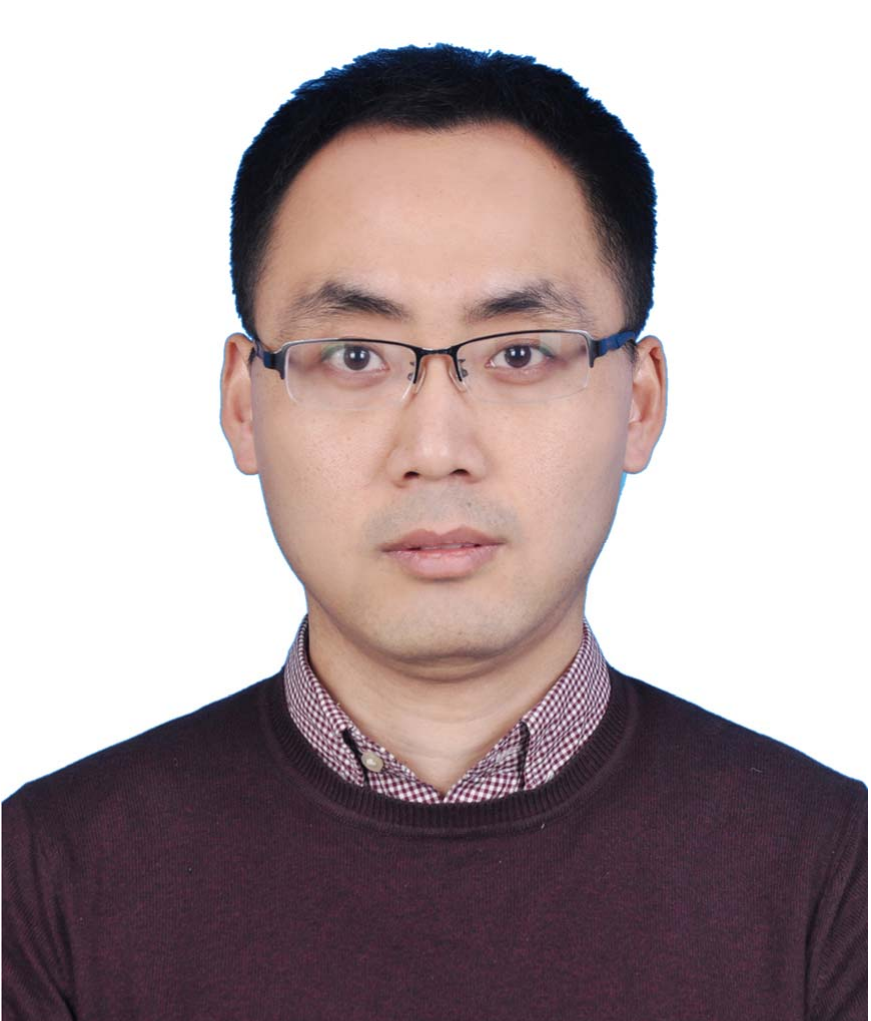}}]{Sanping Zhou}
	received his PhD. degree from Xi'an Jiaotong University, Xi'an, China, in 2020. From 2018 to 2019, he was a Visiting Ph.D. Student with Robotics Institute, Carnegie Mellon University.  He is currently an Associate Professor with the Institute of Artificial Intelligence and Robotics at Xi'an Jiaotong University.  His research interests include machine learning, deep learning and computer vision, with a focus on medical image segmentation, person re-identification, salient object detection, image classification and visual tracking.
\end{IEEEbiography}
\vspace{-30pt} 
\begin{IEEEbiography}[{\includegraphics[width=1.0in,height=1.25in,clip]{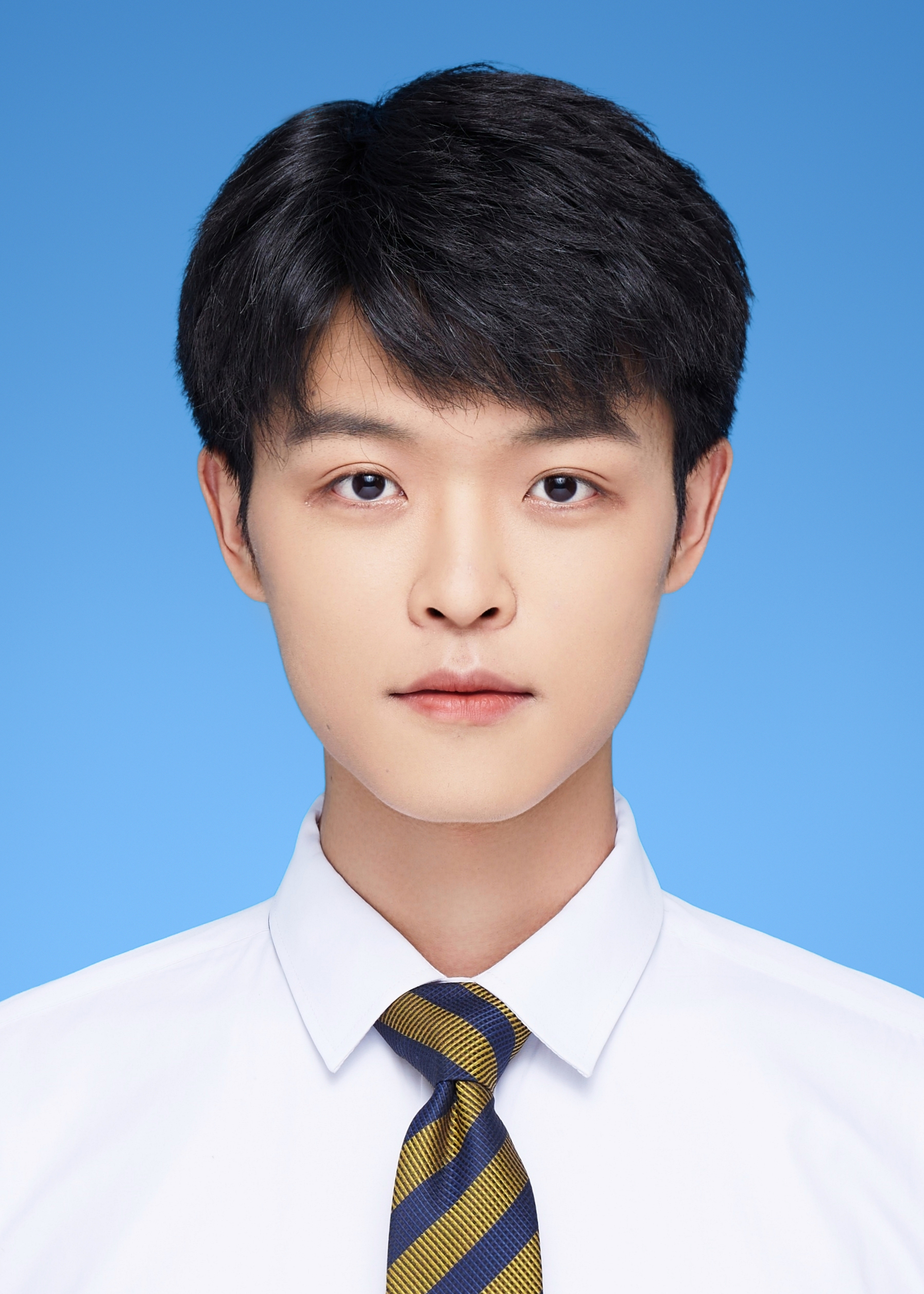}}]{Zheng Qin}
received the B.S. degree in robotic engineering from the Harbin Institute of Technology, China, in 2021. He is currently working toward the PdD. degree in artificial intelligence from Xi'an Jiaotong University. His research interests include computer vision and multi-object tracking.
\end{IEEEbiography}
\vspace{-30pt} 
\begin{IEEEbiography}[{\includegraphics[width=1in,height=1.25in,clip,keepaspectratio]{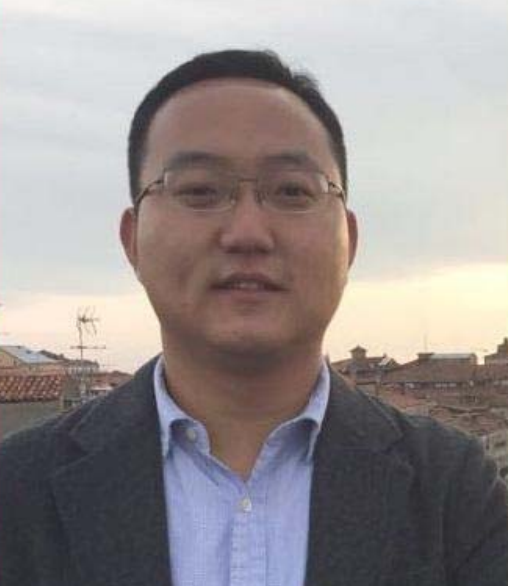}}]{Le Wang}
	(Senior Member, IEEE) received the B.S. and Ph.D. degrees in Control Science and Engineering from Xi'an Jiaotong University, Xi'an, China, in 2008 and 2014, respectively. From 2013 to 2014, he was a visiting Ph.D. student with Stevens Institute of Technology, Hoboken, New Jersey, USA. From 2016 to 2017, he was a visiting scholar with Northwestern University, Evanston, Illinois, USA. He is currently a Professor with the Institute of Artificial Intelligence and Robotics of Xi'an Jiaotong University, Xi'an, China. His research interests include computer vision, pattern recognition, and machine learning.
\end{IEEEbiography}
\vspace{-30pt} 
\begin{IEEEbiography}[{\includegraphics[width=1.0in,height=1.2in,clip,keepaspectratio]{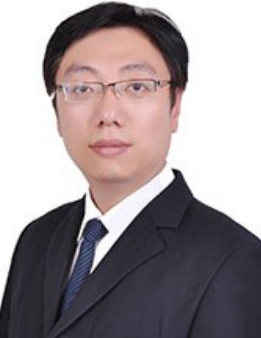}}]{Jinjun Wang}
received the B.E. and M.E. degrees from the Huazhong University of Science and Technology, China, in 2000 and 2003, respectively. He received the Ph.D. degree from Nanyang Technological University, Singapore, in 2006. From 2006 to 2009, he was with NEC Laboratories America, Inc., as a Research Scientist, and Epson Research and Development, Inc., as a Senior Research Scientist, from 2010 to 2013. He is currently a Professor with Xi'an Jiaotong University. His research interests include pattern classification, image/video enhancement and editing, content-based image/video annotation and retrieval, semantic event detection, etc.
\end{IEEEbiography}
\vspace{-30pt} 
\begin{IEEEbiography}[{\includegraphics[width=1.0in,height=1.2in,clip]{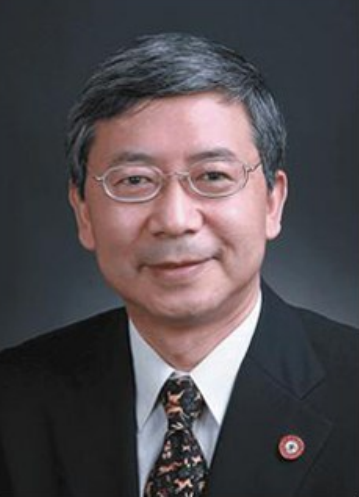}}]{Nanning Zheng}
	(SM93-F06) graduated from the Department of Electrical Engineering, Xian Jiaotong University, Xian, China, in 1975, and received the M.S. degree in information and control engineering from Xian Jiaotong University in 1981 and the Ph.D. degree in electrical engineering from Keio University, Yokohama, Japan, in 1985. He jointed Xian Jiaotong University in 1975, and he is currently a Professor and the Director of the Institute of Artificial Intelligence and Robotics, Xian Jiaotong University. His research interests include computer vision, pattern recognition and image processing, and hardware implementation of intelligent systems. Dr. Zheng became a member of the Chinese Academy of Engineering in 1999, and he is the Chinese Representative on the Governing Board of the International Association for Pattern Recognition.
\end{IEEEbiography}
\end{document}